\documentclass{article}
\usepackage[preprint]{neurips_2026}

\usepackage[utf8]{inputenc}
\usepackage[T1]{fontenc}
\usepackage{hyperref}
\usepackage{url}
\usepackage{booktabs}
\usepackage{amsfonts}
\usepackage{amsmath}
\usepackage{amssymb}
\usepackage{mathtools}
\usepackage{amsthm}
\usepackage{nicefrac}
\usepackage{microtype}
\usepackage{xcolor}
\usepackage{graphicx}
\usepackage{multirow}
\usepackage{enumitem}
\usepackage[capitalize,noabbrev]{cleveref}

\newcommand{\shifamind}{\textsc{ShifaMind}}
\newcommand{\relu}{\mathrm{ReLU}}

\theoremstyle{plain}

\title{\shifamind{}: A Multiplicative Concept Bottleneck for Interpretable ICD-10 Coding}

\author{
   Mohammed Sameer Syed, Xuan Lu \\
   College of Information Science, University of Arizona, Tucson, AZ 85721\\
   \texttt{\{mohammedsameer, luxuan\}@arizona.edu}
}

\begin{document}

\maketitle

\begin{abstract}

Automated ICD-10 coding from clinical discharge summaries requires models that are both accurate on long-tailed multi-label classification tasks and interpretable to clinicians. Concept Bottleneck Models (CBMs) offer a principled framework for interpretability by routing predictions through human-interpretable concepts, but this transparency often comes at a cost: compressing rich clinical text representations into a narrow concept layer can restrict gradient flow and limit predictive capacity.
We present \shifamind{}, a concept-grounded architecture built around a Multiplicative Concept Bottleneck (MCB), which changes the \emph{form}, rather than the \emph{width}, of the bottleneck. Instead of projecting through a narrow concept layer, \shifamind{} uses a learned multiplicative gate over a concept-grounded representation while retaining a scalar concept interface for inspection. On MIMIC-IV top-50 ICD-10 coding, \shifamind{} achieves performance competitive with LAAT, the strongest baseline, across F1, AUC, and ranking metrics, while outperforming five additional ICD-coding baselines and providing concept-mediated explanations. Its substantial gains over a capacity-matched Vanilla CBM in both predictive performance and interpretability-oriented metrics highlight the importance of the bottleneck design. 

\end{abstract}

\section{Introduction}
\label{sec:intro}

Clinical AI systems that assign ICD-10 diagnosis codes from unstructured discharge summaries must satisfy two competing requirements: strong performance on a long-tailed multi-label classification task, and explanations grounded in medical concepts that clinicians can inspect and verify~\cite{rudin2019stop}. Existing approaches typically address these requirements asymmetrically. Attention-based ICD coders such as CAML~\cite{mullenbach2018caml} and LAAT~\cite{vu2020laat}, along with pretrained-language-model variants~\cite{huang2022plmicd,yang2022kept,zhang2025gkiicd}, achieve strong accuracy but provide only token-level attention as explanation, which is not a reliable proxy for model reasoning~\cite{jacovi2020faithfulness}. Post-hoc explanation methods, including feature-attribution methods~\cite{ribeiro2016lime,lundberg2017shap}, share this limitation: they explain a fixed model without constraining how predictions are computed. In contrast, inherently interpretable approaches such as prototype networks~\cite{chen2019protopnet} and concept-based models~\cite{koh2020cbm,kim2018tcav} tie explanations directly to the prediction process. 

Among these, \emph{Concept Bottleneck Models (CBMs)}~\cite{koh2020cbm} are a natural fit for the clinical setting: they predict a vector of human-readable medical concepts (e.g., \emph{fever}, \emph{hypotension}, \emph{anticoagulation}) and use only that vector to predict diagnoses, so each decision can be audited by inspecting which concepts the model estimated to be present. A standard CBM~\cite{koh2020cbm} maps an input note to scalar concept activations, one per concept, and predicts diagnoses from these activations alone. This scalar concept interface enables direct inspection but introduces a capacity-interpretability trade-off~\cite{espinosa2022cem,mahinpei2021leakage}. In multi-label, long-tailed ICD coding, projecting a high-dimensional contextual representation through a narrow sigmoid concept layer can limit gradient flow and predictive capacity.

Recent CBM variants address this trade-off by widening the bottleneck. 
Concept Embedding Models (CEM)~\cite{espinosa2022cem} replace each scalar activation with a $k$-dimensional embedding, and Deep Concept Reasoners (DCR)~\cite{barbiero2023dcr} compose concepts via differentiable logical rules. While these approaches improve predictive performance, they no longer maintain a directly inspectable scalar interface. This suggests that the limitation lies not only in the presence of a bottleneck, but in how the bottleneck is structured. 
We take a complementary approach: rather than widening the bottleneck, we change its \emph{form}. We retain the scalar concept interface, replace the additive concept-to-diagnosis projection with a learned multiplicative gate over a concept-grounded representation, and use the high-dimensional representation to preserve predictive capacity while enforcing concept-mediated prediction.

We instantiate this idea in \shifamind{}, an ICD-10 coding architecture built around a Multiplicative Concept Bottleneck (MCB). 
We evaluate \shifamind{} on MIMIC-IV top-50 ICD-10 coding, considering both diagnostic performance and concept-level interpretability. \shifamind{} outperforms most ICD-coding baselines and achieves competitive performance with the strongest baseline, LAAT~\cite{vu2020laat}, across F1, AUC, and ranking metrics, including a Macro-F1 of 0.712. To isolate the effect of bottleneck form, we further compare \shifamind{} with a capacity-matched Vanilla CBM. Beyond the substantial gap in predictive performance (Macro-F1 $0.712$ vs.\ $0.164$), \shifamind{} achieves higher interpretability scores across Concept-Supported True Positive Rate (CSTPR; $0.704$ vs.\ $0.147$), Concept Influence Magnitude (CIM; $1.314$ vs.\ $0.645$), and Concept-Conditioned Recall (CCR; $0.836$ vs.\ $0.361$). Our results suggest that multiplicative bottlenecks are a promising architectural pattern for building concept-mediated models that balance predictive performance with inspectable decision pathways in high-stakes multi-label settings.

\section{Related Work}
\label{sec:related}

\textbf{ICD Coding.} The International Classification of Diseases (ICD), maintained by the World Health Organization, provides a standardized vocabulary for encoding diagnoses, symptoms, and procedures from clinical documentation. Assigning ICD codes to admission records underpins hospital billing, epidemiological research, and quality measurement, and is typically framed in clinical NLP as a multi-label classification problem over discharge summaries with a long-tailed label distribution. CAML \cite{mullenbach2018caml} introduced per-label attention over convolutional features, and LAAT \cite{vu2020laat} combined BiLSTM encoding with label-aware attention and remains a strong baseline on MIMIC-IV top-50. Pretrained-language-model approaches include PLM-ICD \cite{huang2022plmicd}, which chunks long notes for a biomedical RoBERTa backbone; KEPT \cite{yang2022kept}, a Longformer-based knowledge-injected prompt model; and GKI-ICD \cite{zhang2025gkiicd}, a general knowledge-injection framework combining description, synonym, and hierarchy supervision with R-Drop consistency. We evaluate all five as baselines. These models provide at best attention-weight explanations, which are not guaranteed to be faithful to the underlying computation \cite{jacovi2020faithfulness}.

\textbf{Concept Bottleneck Models.} Concept Bottleneck Models~\cite{koh2020cbm} constrain predictions to flow through an intermediate layer of human-interpretable concepts, making the concept layer the unit of audit. 
Two limitations of standard additive CBMs have been studied in prior work. The first is \emph{concept leakage}, where the concept layer encodes task-relevant information beyond its intended semantics, undermining faithfulness~\cite{mahinpei2021leakage,havasi2022ccm,yeh2020conceptshap}. 
Prior work addresses this either by introducing auxiliary pathways for non-concept information~\cite{havasi2022ccm} or by enforcing disentanglement constraints~\cite{marconato2022glancenets}. 
In contrast, \shifamind{} enforces a no-bypass architecture in which predictions depend only on concept-grounded representations. 

The second limitation is the capacity-interpretability trade-off in expressive multi-label settings. \shifamind{} occupies one point in this design space; two recent variants occupy different points. Concept Embedding Models (CEM)~\cite{espinosa2022cem} widen the bottleneck by replacing scalar concept activations with vector embeddings, improving capacity at the cost of a directly inspectable interface. Deep Concept Reasoners (DCR)~\cite{barbiero2023dcr} compose concepts via differentiable logical rules, but have been demonstrated primarily in smaller-scale settings for single-label tasks.

In contrast, \shifamind{} retains the scalar concept interface while recovering capacity through a multiplicative gate over a concept-grounded representation. 
Because our interpretability metrics (\Cref{sec:xai}) rely on scalar concept-presence indicators, they are not directly applicable to embedding-based models without modification. We therefore compare against a capacity-matched Vanilla CBM to isolate the effect of the bottleneck form, and against state-of-the-art ICD coding models to establish predictive performance.

\section{Method}
\label{sec:method}

Concept bottleneck models provide an interpretable interface by predicting outcomes from human-interpretable concepts. However, standard CBMs impose a narrow scalar bottleneck, compressing rich contextual representations into low-dimensional concept activations. While this enables inspection, it limits representational capacity and weakens gradient flow, leading to degraded performance in complex multi-label settings such as ICD-10 coding. Recent variants address this limitation by widening the bottleneck, for example by replacing scalar concepts with embeddings, but this weakens the directly inspectable concept interface.

\shifamind{} is designed around a different principle: \emph{change the form of the bottleneck, not its width}. Given a discharge summary $x$, \shifamind{} predicts diagnosis logits $\hat{\boldsymbol{\ell}} \in \mathbb{R}^L$ over $L$ ICD-10 codes by routing information through a high-dimensional \emph{concept-grounded representation}. This representation is constructed by a cross-attention module in which $C$ learnable queries, one initialized per named concept, attend to the input note. A multiplicative bottleneck then enforces that diagnosis predictions depend only on this concept-grounded representation, preserving encoder-level capacity while maintaining a concept-mediated prediction pathway.

\begin{figure}[t]
    \centering
    \includegraphics[width=1\linewidth]{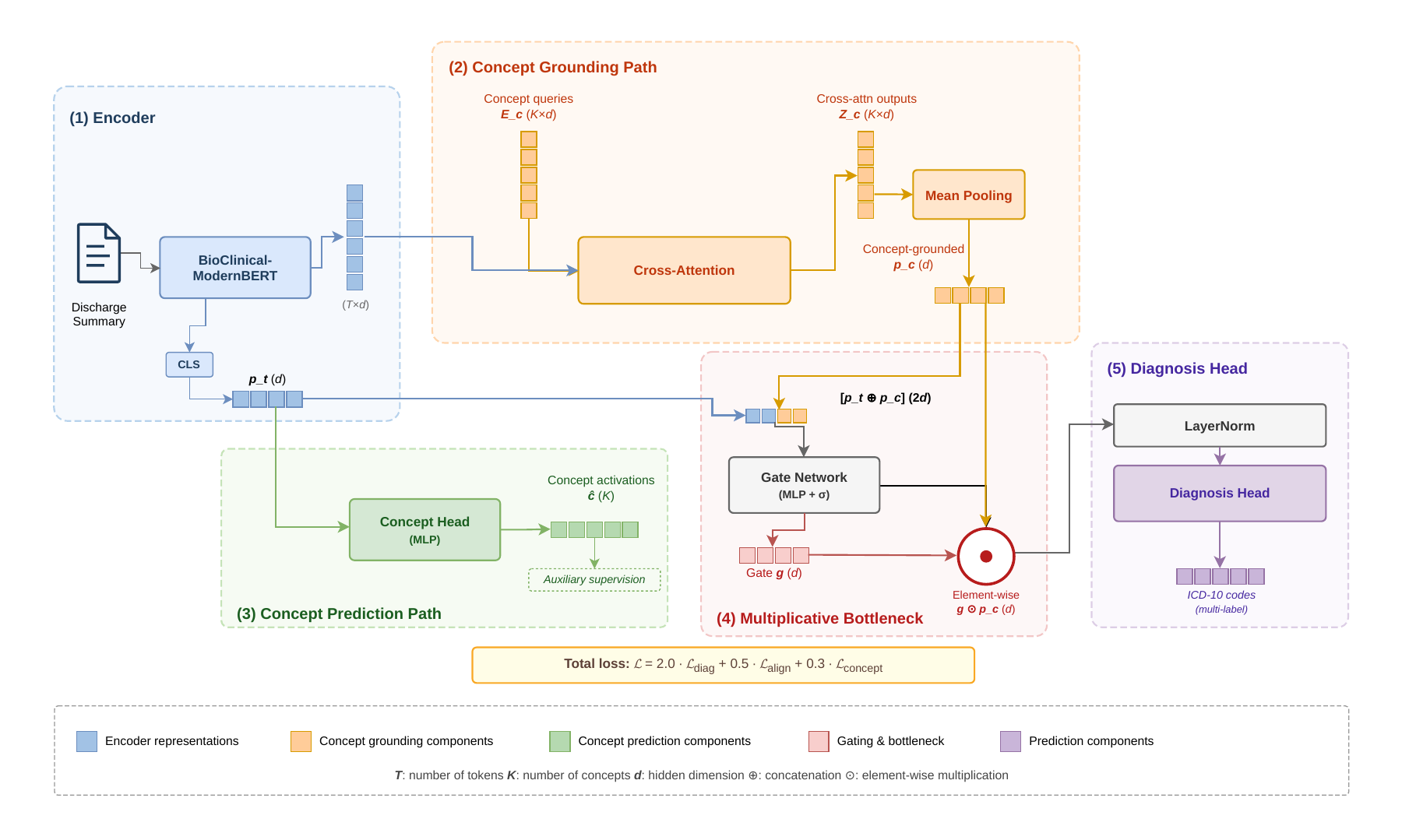}
    \caption{\shifamind{} architecture. A discharge summary is encoded into token and pooled representations. Learnable concept queries produce a concept-grounded representation, while an auxiliary concept head predicts inspectable concept activations (not used for diagnosis). A gated bottleneck modulates the concept-grounded representation before the diagnosis head predicts ICD-10 codes.}
    \label{fig:architecture}
\end{figure}

The model consists of a long-context encoder that produces token representations, a concept grounding module that extracts concept-specific evidence, a multiplicative bottleneck that constrains prediction to concept-grounded features, and a diagnosis head. We additionally train a separate concept head to produce scalar concept activations for clinician inspection; this head is not used in the diagnosis pathway. \Cref{fig:architecture} provides an overview of the full architecture.

\textbf{Long-Context Encoding.} We use BioClinical ModernBERT-base~\cite{bioclinicalmodernbert}, a clinical adaptation of ModernBERT~\cite{warner2025modernbert} that supports context windows up to 8{,}192 tokens. Given an input document $x$, the encoder produces token-level representations
\[
\mathbf{H} \in \mathbb{R}^{n \times h}, \quad h = 768,
\]
where $n$ is the number of tokens. We use the CLS token as a global summary:
\[
\mathbf{p_t} = \mathbf{H}[0, :] \in \mathbb{R}^h.
\]

\textbf{Concept Grounding.} We represent $C$ clinical concepts using learnable query embeddings
\[
\mathbf{E}_c \in \mathbb{R}^{C \times h}.
\]
Each concept attends to the token representations via multi-head attention:
\begin{equation}
\mathbf{Z}_c = \mathrm{MultiHeadAttn}(Q=\mathbf{E}_c, K=\mathbf{H}, V=\mathbf{H}) \in \mathbb{R}^{C \times h}.
\end{equation}

Each row $\mathbf{Z}_c[k,:]$ captures the evidence for concept $k$ in the document. We aggregate across concepts to obtain a concept-grounded representation:
\begin{equation}
\mathbf{p_c} = \frac{1}{C} \sum_{k=1}^{C} \mathbf{Z}_c[k,:] \in \mathbb{R}^h.
\end{equation}

This representation retains the full dimensionality of the encoder while restricting information to concept-relevant subspaces.

\textbf{Multiplicative Concept Bottleneck.} We constrain predictions to depend only on concept-grounded features via a multiplicative gating mechanism. First, we compute a gate conditioned on both the encoder summary $\mathbf{p_t}$ and the concept representation $\mathbf{p_c}$:
\begin{equation}
\mathbf{g} = \sigma\!\left(\mathbf{W}_2 \, \mathrm{ReLU}(\mathbf{W}_1 [\mathbf{p_t}; \mathbf{p_c}] + \mathbf{b}_1) + \mathbf{b}_2 \right) \in [0,1]^h.
\end{equation}

We then apply the gate element-wise to the concept representation:
\begin{equation}
\mathbf{z} = \mathrm{LayerNorm}(\mathbf{g} \odot \mathbf{p_c}),
\end{equation}
and compute diagnosis logits:
\begin{equation}
\hat{\boldsymbol{\ell}} = \mathbf{W}_d \mathbf{z} + \mathbf{b}_d.
\end{equation}

Intuitively, the model first extracts concept-specific evidence and then uses the encoder representation only to modulate which concept dimensions are relevant for prediction.

\textbf{\textit{No-bypass property.}} 
There is no direct path from $\mathbf{p_t}$ or $\mathbf{H}$ to the prediction head; the diagnosis logits depend on $\mathbf{p_t}$ only as a multiplicative modulator of $\mathbf{p_c}$ via the gate. In particular, if $\mathbf{p_c} = \mathbf{0}$, then $\hat{\boldsymbol{\ell}}$ reduces to a constant.

\textbf{Concept Supervision via NegEx.}
To enable interpretability, we train a separate concept head that produces scalar concept activations:
\begin{equation}
\hat{\mathbf{c}} = \sigma(\mathbf{W}_c \mathbf{p_t} + \mathbf{b}_c) \in [0,1]^C.
\end{equation}

We derive pseudo-labels using a negation-aware rule-based system based on NegEx~\cite{chapman2001negex}, which identifies negation triggers (e.g., \textit{no}, \textit{denies}, \textit{without}) and marks concepts as positive only when they appear in non-negated contexts. Implementation details are in \Cref{app:negex}. These pseudo-labels provide weak supervision for the concept head but are not used as prediction targets. The diagnosis pathway depends solely on the concept-grounded representation $\mathbf{p_c}$.

\textbf{Training Objective.}
We optimize a joint objective:
\begin{equation}
\mathcal{L} =
\lambda_{\text{diag}} \mathcal{L}_{\text{diag}} +
\lambda_{\text{align}} \mathcal{L}_{\text{align}} +
\lambda_{\text{concept}} \mathcal{L}_{\text{concept}}.
\end{equation}

$\mathcal{L}_{\text{diag}}$ is a focal loss~\cite{lin2017focalloss} over ICD-10 labels to address class imbalance.  
$\mathcal{L}_{\text{concept}}$ is a binary cross-entropy loss between $\hat{\mathbf{c}}$ and the pseudo-labels.  
$\mathcal{L}_{\text{align}}$ is a cosine similarity regularizer between $\mathbf{p_t}$ and $\mathbf{p_c}$ that stabilizes training.
Unless otherwise specified, we set $(\lambda_{\text{diag}}, \lambda_{\text{align}}, \lambda_{\text{concept}}) = (2.0, 0.5, 0.3)$. Optimization details are provided in \Cref{app:training}.

\section{Experimental Setup}
\label{sec:experiments}

\subsection{Dataset}
\label{sec:dataset}

We use MIMIC-IV v3.1 \cite{johnson2023mimiciv}, linking ICD-10 diagnoses from the hospital admissions table to discharge summaries from MIMIC-IV-Note \cite{johnson2023mimicivnote}. Following the standard top-50 ICD-coding benchmark construction~\cite{mullenbach2018caml,vu2020laat,edin2023automated}, we select $L=50$ ICD-10 codes by admission count. We exclude three non-clinical external-cause codes (Y92230 and Y929, ``place of occurrence''; Z20822, ``contact with COVID-19 exposure'') that lack reliable textual evidence in discharge notes and yielded near-zero $F_1$ scores in preliminary experiments. We replace them with the next most frequent clinically grounded codes: E876 (hypokalemia), E875 (hyperkalemia), and D72829 (elevated white blood cell count). After filtering to admissions with both a discharge note and at least one selected code, the final dataset contains 113{,}918 admissions, split 70/15/15 into 79{,}742 training, 17{,}088 validation, and 17{,}088 test admissions using seed 42. The full code list and per-code training counts are provided in \Cref{app:icd_codes}.

\textbf{Concept Vocabulary.} 
We construct the concept vocabulary to cover all target ICD-10 codes: each code maps to at least one clinical concept. The vocabulary includes symptoms (\emph{fever}, \emph{dyspnea}, \emph{chest}), vital-sign and laboratory abnormalities (\emph{hypotension}, \emph{hyponatremia}, \emph{acidosis}), organ-system categories (\emph{cardiac}, \emph{pulmonary}, \emph{renal}), medications and interventions (\emph{insulin}, \emph{anticoagulation}, \emph{CPAP}), and condition-level terms (\emph{diabetes}, \emph{copd}, \emph{fibrillation}, \emph{palliative}). To ensure adequate empirical support, we retain concepts that appear in at least $1\%$ of a 10{,}000-note sample from the training set, resulting in $C=160$ concepts. Under NegEx supervision, an average of 24.7 concepts are activated per note, indicating that notes are represented by multiple clinical concepts rather than isolated sparse signals. The full vocabulary is provided in \Cref{app:concept_list}.

\subsection{Baselines}
\label{sec:baselines}

We compare \shifamind{} against six representative baselines covering major families of ICD coding models: convolutional attention, recurrent label attention, pretrained language models (PLMs) with chunking, long-context PLMs with knowledge injection, PLMs with cross-attention decoders, and concept bottleneck models. These baselines span the main design axes of ICD coding models: backbone architecture, context handling, knowledge integration, and interpretability. All baselines are trained and evaluated on the same MIMIC-IV top-50 split using a shared global thresholding protocol (\Cref{sec:threshold}). Implementation details follow the original papers; full hyperparameters are provided in \Cref{app:baseline_details}.

\textbf{CAML}~\cite{mullenbach2018caml} is a convolutional attention model that applies a 1D CNN over Word2Vec embeddings and aggregates features with per-label attention to produce diagnosis logits.

\textbf{LAAT}~\cite{vu2020laat} replaces the convolutional backbone with a BiLSTM encoder followed by label-aware attention. It is a strong non-PLM baseline.

\textbf{PLM-ICD}~\cite{huang2022plmicd} uses a pretrained transformer encoder with chunking to handle long documents. We adopt \texttt{biomed\_roberta\_base}, splitting each note into fixed-length segments and applying a label-aware attention head over concatenated representations.

\textbf{KEPT}~\cite{yang2022kept} extends PLM-based coding to long contexts using a Longformer backbone with knowledge-enhanced pretraining. It encodes both ICD descriptions and discharge notes, using masked positions associated with each code to produce predictions.

\textbf{GKI-ICD}~\cite{zhang2025gkiicd} combines a chunked PLM encoder with a cross-attention decoder over code-specific queries. It incorporates auxiliary supervision from guideline-based knowledge during training.

\textbf{Vanilla CBM}~\cite{koh2020cbm} is the key interpretability baseline. We implement a capacity-matched version using the same BioClinical-ModernBERT-base encoder, context length, optimizer, and loss as \shifamind{}. Predictions are made via a standard additive bottleneck:
\[
\hat{\mathbf{c}} = \sigma(\mathbf{W}_c \mathbf{p_t} + \mathbf{b}_c), \quad
\hat{\boldsymbol{\ell}} = \mathbf{W}_d \hat{\mathbf{c}} + \mathbf{b}_d,
\]
where $\hat{\boldsymbol{\ell}}$ denotes diagnosis logits. This is a strict CBM: diagnosis predictions depend only on the scalar concept activations $\hat{\mathbf{c}}$, with no direct path from the encoder representation $\mathbf{p_t}$ to the prediction head. Thus, performance differences more directly reflect differences in bottleneck design.

\subsection{Evaluation Protocol}
\label{sec:threshold}

We use a fixed global threshold of $\tau = 0.5$ (applied to sigmoid outputs) for all models, without validation tuning. Prior work often tunes a per-model threshold to optimize F1~\cite{mullenbach2018caml,vu2020laat,huang2022plmicd}, but this introduces an additional degree of freedom that depends on model calibration. Threshold sensitivity has also been documented as a reproducibility concern in ICD coding~\cite{edin2023automated}. Using a shared threshold enables a more consistent comparison across models under a common decision rule. 

To complement threshold-based metrics, in addition to Macro and Micro F1, we report threshold-independent measures: Macro and Micro AUC-ROC, as well as Precision@$K$ and Recall@$K$ for $K \in \{5, 8, 15\}$. These metrics capture ranking performance independent of threshold choice.

\subsection{Training Dynamics}
\begin{figure}[t]
    \centering
    \includegraphics[width=0.5\linewidth]{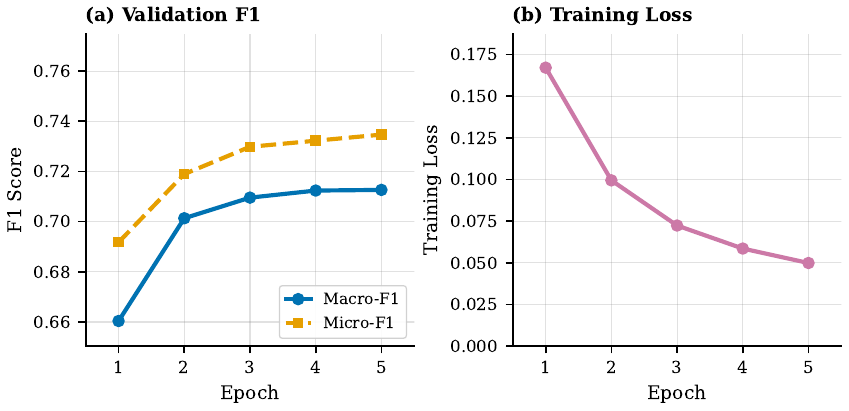}
    \caption{\shifamind{} training dynamics. Left: validation Macro-F1 and Micro-F1 across five epochs. Right: total training loss.}
    \label{fig:training}
\end{figure}

\Cref{fig:training} shows the training curves for \shifamind{}. Validation Macro-F1 increases from 0.660 at epoch 1 to 0.712 at epoch 5, while the joint training loss decreases monotonically from 0.167 to 0.050. The concept and alignment losses also continue to decrease through epoch 5, suggesting stable optimization within the five-epoch training budget.

\section{Results}
\label{sec:results}

\subsection{Diagnostic Performance}
\label{subsec:diagnose}

\Cref{tab:main_results} reports test-set performance across all seven models. \shifamind{} achieves a Macro-F1 of $0.712$, slightly outperforming LAAT, although the difference is not statistically significant. All other baselines perform significantly worse (see \Cref{app:bootstrap} for bootstrap tests). The capacity-matched Vanilla CBM performs substantially worse (Macro-F1 $0.164$), highlighting the importance of the bottleneck form; we analyze this comparison further in \Cref{sec:ablation}.

We focus on Macro-F1 as the primary metric, as it weights all $L=50$ ICD-10 codes equally and is therefore appropriate for the long-tailed label distribution (\Cref{fig:code_distribution}), where rare but clinically important diagnoses should not be ignored. 
To assess performance across frequency regimes, we partition the 50 codes into HEAD, MID, and TAIL groups (16/16/18). \shifamind{} achieves the best Macro-F1 on HEAD and MID codes and remains competitive on TAIL codes, where LAAT performs slightly better. This suggests that the overall gain is not driven only by frequent labels. Full per-bin results are reported in \Cref{tab:longtail}.

On threshold-independent ranking metrics (AUC and Precision@$K$/Recall@$K$), \shifamind{} ranks second, slightly behind LAAT, but significantly outperforms the remaining baselines ($p < 0.01$; \Cref{app:bootstrap}).

\begin{table}[t]
\centering
\footnotesize
\caption{Main results on MIMIC-IV top-50 ICD-10 coding. \textbf{Bold} indicates best performance and \underline{underline} indicates second best.} 
\label{tab:main_results}
\begin{tabular}{lcccccccccccc}
\toprule
\textbf{Model}
  & \textbf{Mac-F1} & \textbf{Mic-F1}
  & \textbf{Mac-AUC} & \textbf{Mic-AUC}
  & \textbf{P@5} & \textbf{P@8} & \textbf{P@15} & \textbf{R@5} & \textbf{R@8} & \textbf{R@15}
   \\
\midrule
CAML \cite{mullenbach2018caml}        & 0.674           & 0.716           & 0.939           & 0.954           & 0.633           & 0.508           & 0.325           & 0.698           & 0.833           & 0.942 \\
LAAT \cite{vu2020laat}                & \underline{0.711} & \textbf{0.746} & \textbf{0.947} & \textbf{0.961} & \textbf{0.654} & \textbf{0.521} & \textbf{0.329} & \textbf{0.718} & \textbf{0.850} & \textbf{0.950} \\
PLM-ICD \cite{huang2022plmicd}        & 0.650           & 0.699           & 0.929           & 0.947           & 0.623           & 0.495           & 0.317           & 0.686           & 0.814           & 0.925 \\
KEPT \cite{yang2022kept}              & 0.687           & 0.727           & 0.941           & 0.957           & 0.643           & 0.508           & 0.323           & 0.706           & 0.833           & 0.939 \\
GKI-ICD \cite{zhang2025gkiicd}        & 0.649           & 0.700           & 0.930           & 0.949           & 0.626           & 0.496           & 0.318           & 0.689           & 0.816           & 0.928 \\
Vanilla CBM \cite{koh2020cbm}         & 0.164           & 0.372           & 0.699           & 0.779           & 0.374           & 0.317           & 0.235           & 0.396           & 0.528           & 0.708 \\
\midrule
\shifamind{}                   & \textbf{0.712}  & \underline{0.735} & \underline{0.942} & \underline{0.959} & \underline{0.647} & \underline{0.517} & \underline{0.327} & \underline{0.711} & \underline{0.845} & \underline{0.947} \\
\bottomrule
\end{tabular}
\end{table}

\subsection{Ablation Study}
\label{sec:ablation}

We ablate two components of \shifamind{} while holding all other factors fixed (backbone, data split, training schedule, and loss weights except the one being ablated): (i) \textbf{w/o alignment loss}, setting $\lambda_{\mathrm{align}} = 0$; and (ii) \textbf{w/o cross-attention}, replacing the concept-grounded representation $\mathbf{p_c}$ with the CLS representation $\mathbf{p_t}$ so that the multiplicative gate operates without concept grounding. \Cref{tab:ablation} reports the results. For reference, we also include the capacity-matched Vanilla CBM (additive bottleneck on $\mathbf{p_t}$), which removes both the multiplicative gate and the concept-grounded representation.

Removing cross-attention leads to a noticeable drop in performance (Macro-F1 $0.712 \rightarrow 0.689$; Macro-AUC $0.942 \rightarrow 0.936$), indicating that concept grounding contributes beyond what can be recovered from the encoder representation alone. In contrast, removing the alignment loss results in only a small change ($+0.003$ Macro-F1), suggesting that it has limited impact on final performance under this setup. We retain it in the full model as it stabilizes training in early stages.
Compared to the Vanilla CBM (Macro-F1 $0.164$), both gated variants perform substantially better, highlighting the importance of the bottleneck form. We analyze this comparison further in \Cref{sec:xai}.

\begin{table}[t]
\centering
\small
\caption{Ablation of \shifamind{} on the MIMIC-IV top-50 test set. All variants share the same backbone and training setup.}
\label{tab:ablation}
\begin{tabular}{lccc}
\toprule
\textbf{Configuration} & \textbf{Macro-F1} & \textbf{Macro-AUC} & \textbf{$\Delta$ F1} \\
\midrule
\shifamind{} (full)                                  & 0.712 & 0.942 & --- \\
\quad w/o alignment loss ($\lambda_{\mathrm{align}}=0$) & 0.715 & 0.945 & +0.003 \\
\quad w/o cross-attention ($\mathbf{p_c} \leftarrow \mathbf{p_t}$) & 0.689 & 0.936 & $-0.023$ \\
\midrule

Vanilla CBM (additive bottleneck on $\mathbf{p_t}$)  & 0.164 & 0.699 & $-0.548$ \\
\bottomrule
\end{tabular}
\end{table}

\subsection{Interpretability Evaluation}
\label{sec:xai}

Beyond diagnostic accuracy, we evaluate whether the learned concept pathway provides meaningful and faithful concept-level explanations. Most ICD-coding baselines in our comparison do not provide intrinsic concept-level interpretability: PLM-ICD, KEPT, and GKI-ICD do not expose an interpretable intermediate concept layer, while CAML and LAAT provide attention weights only. We therefore focus the interpretability comparison on the capacity-matched Vanilla CBM, which shares the same goal of concept-level prediction. Following the view that interpretability claims should be evaluated through testable behavioral properties rather than architectural assumptions~\cite{jacovi2020faithfulness,doshivelez2017rigorousML}, we compare \shifamind{} with the capacity-matched Vanilla CBM using three complementary metrics: whether correct predictions are supported by relevant concepts (CSTPR), whether the concept-grounded representation influences predictions (CIM), and whether diagnoses are correctly identified when relevant clinical concepts are present (CCR). Because both models share the same backbone, context length, and training setup, this comparison isolates the effect of the bottleneck form. We report bootstrap 95\% confidence intervals over $1{,}000$ test-set resamples.

\textbf{Metric 1: CSTPR (Concept-Supported True Positive Rate).} Following the faithfulness framework of Jacovi and Goldberg \cite{jacovi2020faithfulness}, CSTPR asks: \emph{of all truly positive diagnoses, how many are both  correctly predicted and supported by at least one correctly predicted relevant concept?} For each label $j$, $\mathrm{TopC}(j)$ is the set of five concepts with highest Pearson correlation with $j$ on the training set. We define
\[
\mathrm{CSTPR}_j = \frac{\#\{i : y_{ij}=1,\; \hat{y}_{ij}=1,\; \exists c \in \mathrm{TopC}(j) \text{ s.t.\ } \hat{c}_{ic} > 0.5 \text{ and } \tilde{c}_{ic}=1\}}{\#\{i : y_{ij}=1\}},
\]
and report the macro-average over labels.

\textbf{Metric 2: CIM (Concept Influence Magnitude).} Following gradient-based sensitivity analysis \cite{simonyan2014gradients}, CIM measures how strongly the representation driving the diagnosis head influences the output logits. For each concept-diagnosis pair $(c,j)$, we compute 
\[
\mathrm{CIM}_{c,j} = \mathbb{E}_{i \in \mathrm{copos}(c,j)}\left[\|\nabla_{\mathbf{r}^{(i)}}\hat{\ell}_j^{(i)}\|_2\right]
\]
where $\mathrm{copos}(c,j) = \{i : \tilde{c}_{ic}=1,\, y_{ij}=1\}$ and $\mathbf{r}$ denotes the representation input to the diagnosis head ($\mathbf{p_c}$ for \shifamind{}, $\hat{\mathbf{c}}$ for the Vanilla CBM; the literal input to each model's diagnosis head). 
Gradients are computed analytically (\Cref{app:jacobian}).

\textbf{Metric 3: CCR (Concept-Conditioned Recall).} Following Doshi-Velez and Kim \cite{doshivelez2017rigorousML}, CCR evaluates whether the model recovers diagnoses when relevant concepts are present. For each pair $(c,j)$, 
\[
\mathrm{CCR}_{c,j} = P(\hat{y}_j = 1 \mid y_j = 1,\; \tilde{c}_{c} = 1)
\]
i.e., recall of label $j$ restricted to samples where concept $c$ is present.

\begin{figure}[t]
    \centering
    \includegraphics[width=0.7\linewidth]{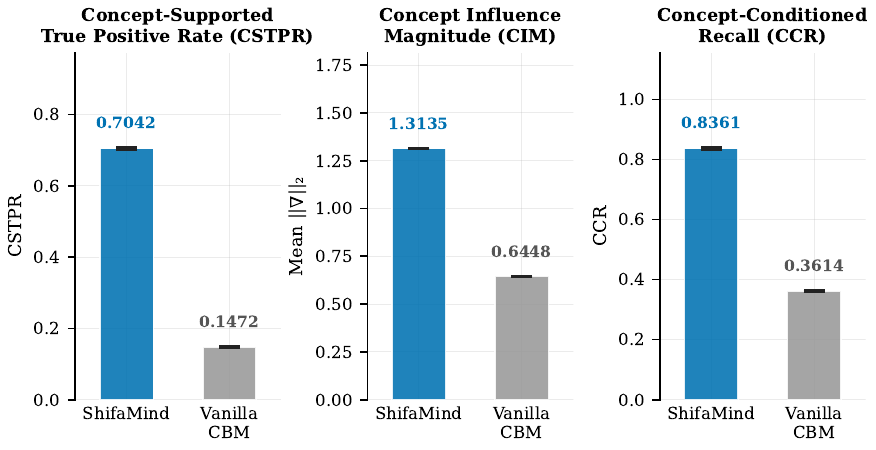}
    \caption{Side-by-side interpretability comparison. \shifamind{} MCB outperforms the capacity-matched Vanilla CBM on all three metrics, with non-overlapping bootstrap 95\% confidence intervals.}
    \label{fig:xai}
\end{figure}

All three metrics are non-negative, and higher values indicate stronger concept-supported predictive behavior. CSTPR and CCR are rates in $[0,1]$, while CIM is a gradient-norm sensitivity measure whose scale is model-dependent and is therefore interpreted comparatively. \Cref{fig:xai} reports the results: \shifamind{} achieves higher CSTPR, CIM, and CCR than the capacity-matched Vanilla CBM, with non-overlapping bootstrap 95\% confidence intervals across all three metrics.

\textbf{Architectural interpretation.}
These differences are consistent with the two models' bottleneck designs. Vanilla CBM compresses the 768-dimensional encoder representation into 160 scalar concept activations before prediction:
\[
\mathbf{p_t} \rightarrow \mathbf{W}_c \rightarrow \sigma \rightarrow \mathbf{W}_d .
\]
This scalar bottleneck limits representational capacity. In contrast, \shifamind{} preserves a 768-dimensional concept-grounded representation and uses the gate as a multiplicative modulator rather than a scalar prediction bottleneck. The gradient magnitude at each model's diagnosis-head input is correspondingly larger for \shifamind{} (CIM ratio $2.0\times$, \Cref{fig:xai}). The CCR results show that this difference also appears behaviorally: when a relevant ground-truth concept is present, \shifamind{} recovers the associated diagnosis more often than Vanilla CBM ($83.6\%$ vs.\ $36.1\%$). 

\textbf{Behavioral test via concept-mask intervention.} As a complementary behavioral test, we mask token spans corresponding to the target diagnosis's TopC concepts and re-run inference. We sample 1{,}000 correctly predicted note--diagnosis pairs at $\tau=0.5$ and compare the drop in the target diagnosis probability with the average drop for other positive diagnoses in the same note. Among the 917 pairs, where TopC concepts appear in the note, the target diagnosis probability drops more than the within-note control by $0.114$ on average (target: $0.134$; control: $0.020$), with a paired bootstrap 95\% CI of $[0.103, 0.127]$. The target drop is larger than the within-note control in 631/917 pairs ($68.8\%$; two-sided binomial sign test $p \ll 0.001$). These results provide additional behavioral evidence that \shifamind{} predictions are sensitive to clinically relevant concept mentions, complementing the structural no-bypass design. Full details of span matching, masking, sampling, and control construction are provided in \Cref{app:concept_mask}.

\section{Discussion}

\subsection{Limitations}
\label{sec:limitations}

\textbf{Single-seed evaluation.} We report results from a single training seed for each configuration due to compute constraints. The closest comparison, \shifamind{} versus LAAT on Macro-F1, is statistically tied at this seed. Multi-seed experiments would clarify whether the small performance gap and the alignment-loss ablation effect ($+0.003$ Macro-F1; \Cref{tab:ablation}) are meaningful or within seed variance.

\textbf{Concept supervision.} We train the concept head using NegEx pseudo-labels rather than expert-annotated concepts. Although the $10.8\%$ negation correction rate suggests that NegEx improves over naive keyword matching, residual label noise remains, especially for concepts with complex semantic scope. Expert annotation would provide a stronger estimate of concept-head quality, but is difficult at the scale of the 79{,}742-note training set. Validation on a smaller expert-labeled subset is an important next step.

\textbf{Vocabulary scaling and code coverage.} The MCB architecture is not tied to fixed values of $L$ or $C$: the diagnosis head $\mathbf{W}_d \in \mathbb{R}^{L \times h}$ and concept-query bank $\mathbf{E}_c \in \mathbb{R}^{C \times h}$ scale linearly with the number of labels and concepts. However, the 160-concept vocabulary was designed for the top-50 ICD setting. Extending to broader ICD subsets would require a larger concept vocabulary, while full ICD-10 coding introduces additional long-tail challenges documented in prior work~\cite{edin2023automated}.

\textbf{Single dataset.} Our evaluation is limited to MIMIC-IV. External validation on additional clinical corpora is needed before deployment.

\textbf{Clinical validation.} CSTPR and CCR assess whether concept activations align with relevant diagnoses, but they do not measure whether clinicians find the explanations useful in practice. A clinician user study is an important direction for future work.

\textbf{Concept-pathway evaluation.} Our evidence for concept mediation combines architectural (no-bypass), statistical (CSTPR/CIM/CCR), and behavioral intervention tests. Each provides only a partial view. In particular, the masking intervention tests whether predictions are sensitive  to TopC concept mentions, but it does not establish that each learned concept query corresponds one-to-one to its initialization label. We therefore interpret \shifamind{}'s queries as a learned concept-grounded representation rather than as independently validated named-concept detectors.

\subsection{Broader Impacts}
\label{sec:broader_impacts}
Automated ICD-10 coding has the potential to reduce clinical documentation burden, improve consistency in hospital billing, and support epidemiological surveillance. Concept-mediated explanations, such as those produced by \shifamind{}, aim to make automated coding decisions more inspectable and auditable. However, the technology also carries risks: automation bias if coders defer uncritically to model outputs, demographic disparities if concept vocabularies or training data underrepresent certain populations or conditions, and downstream errors propagating from coding into billing or research. We position \shifamind{} as a decision-support tool for clinical coders rather than a replacement, and emphasize that external validation on additional clinical corpora is required before any deployment.

\section{Conclusion}

We presented \shifamind{}, an ICD-10 coding architecture built around a Multiplicative Concept Bottleneck (MCB) that predicts diagnoses through a multiplicative gate over a concept-grounded representation. This design addresses the capacity--interpretability trade-off by preserving an inspectable scalar concept interface while retaining the predictive capacity needed for multi-label, long-tailed ICD coding. On the MIMIC-IV top-50 ICD-10 coding task, under a shared global threshold, \shifamind{} achieves competitive performance with the strongest baseline, LAAT, across F1, AUC, and ranking metrics; outperforms the remaining baselines; and provides concept-mediated explanations. A capacity-matched additive CBM performs substantially worse in both predictive performance and interpretability-oriented metrics, highlighting the importance of the bottleneck form. These results suggest that multiplicative bottlenecks may offer a useful architectural pattern for multi-label concept-bottleneck tasks; validation on additional datasets and domains remains future work.

\bibliographystyle{plain}
\bibliography{shifamind}

\newpage
\appendix

\section{ICD-10 Code List and Per-Code Performance}
\label{app:icd_codes}
\Cref{tab:per_code_f1} reports per-code $F_1$ for \shifamind{} on the MIMIC-IV test set, alongside the number of positive admissions per code. $F_1$ ranges from $0.912$ (Z951, cardiac device \emph{in situ}) to $0.457$ (D72829, elevated WBC). Per-code performance broadly tracks admission prevalence but with notable exceptions in both directions: Z951 achieves high $F_1$ at moderate $N = 11{,}267$, whereas Z87891 (history of nicotine dependence) reaches only $F_1 = 0.545$ despite $N = 62{,}803$.
\begin{table}[h]
\centering
\small
\caption{\shifamind{} per-code $F_1$ on the MIMIC-IV test set, global threshold $\tau = 0.5$, sorted by $F_1$. N is the number of positive admissions.}
\label{tab:per_code_f1}
\begin{tabular}{lcr lcr lcr lcr}
\toprule
\textbf{Code} & \textbf{$F_1$} & \textbf{N} & \textbf{Code} & \textbf{$F_1$} & \textbf{N} & \textbf{Code} & \textbf{$F_1$} & \textbf{N} & \textbf{Code} & \textbf{$F_1$} & \textbf{N} \\
\midrule
Z951   & 0.912 & 11{,}267 & J45909 & 0.788 & 20{,}677 & Z7901 & 0.734 & 30{,}956 & E876    & 0.606 & 9{,}735 \\
E039   & 0.898 & 27{,}999 & G4733  & 0.787 & 23{,}933 & I110  & 0.733 & 15{,}423 & E1165   & 0.586 & 10{,}248 \\
I10    & 0.870 & 83{,}773 & I252   & 0.785 & 15{,}275 & D62   & 0.732 & 19{,}356 & G8929   & 0.580 & 14{,}786 \\
I2510  & 0.850 & 41{,}548 & Z66    & 0.780 & 18{,}899 & N189  & 0.723 & 15{,}379 & E872    & 0.570 & 15{,}995 \\
E785   & 0.844 & 84{,}568 & M109   & 0.777 & 11{,}531 & Z8673 & 0.717 & 15{,}126 & K5900   & 0.551 & 12{,}860 \\
Z794   & 0.826 & 27{,}640 & N179   & 0.775 & 35{,}884 & I5032 & 0.691 & 10{,}787 & Z7902   & 0.549 & 19{,}961 \\
K219   & 0.816 & 56{,}155 & Z86718 & 0.766 & 13{,}706 & E871  & 0.687 & 17{,}068 & Z87891  & 0.545 & 62{,}803 \\
Z955   & 0.808 & 14{,}481 & I4891  & 0.762 & 21{,}022 & E875  & 0.655 & 9{,}654  & F17210  & 0.535 & 24{,}106 \\
J449   & 0.803 & 18{,}023 & N400   & 0.755 & 12{,}758 & J9601 & 0.643 & 13{,}339 & D649    & 0.498 & 21{,}137 \\
F329   & 0.799 & 41{,}876 & I130   & 0.754 & 12{,}946 & D509  & 0.628 & 10{,}045 & D696    & 0.489 & 11{,}260 \\
E1122  & 0.799 & 18{,}564 & I129   & 0.749 & 16{,}650 & G4700 & 0.627 & 12{,}134 & D72829  & 0.457 & 9{,}435 \\
N390   & 0.793 & 16{,}696 & I480   & 0.747 & 13{,}399 & E669  & 0.608 & 22{,}004 &   &   &   \\
       &        &          & F419   & 0.747 & 38{,}910 & & & & & & \\
       &        &          & E119   & 0.746 & 26{,}266 & & & & & & \\
       &        &          & Z515   & 0.736 & 9{,}752  & & & & & & \\
\bottomrule
\end{tabular}
\end{table}

\section{Concept Vocabulary}
\label{app:concept_list}

The 160-concept vocabulary used by \shifamind{}, organized into semantic groups, is listed in full below:

\begin{quote}\small\emph{Symptoms/signs:} fever, cough, dyspnea, pain, nausea, vomiting, diarrhea, fatigue, headache, dizziness, weakness, confusion, syncope, chest, abdominal, dysphagia, hemoptysis, hematuria, hematemesis, melena, jaundice, edema, rash, pruritus, weight, anorexia, malaise, wheezing, reflux, constipation, bowel.\\
\emph{Vitals/states:} hypotension, hypertension, tachycardia, bradycardia, tachypnea, hypoxia, hypothermia, shock, altered, lethargic, obtunded.\\
\emph{Organ systems:} cardiac, pulmonary, renal, hepatic, neurologic, gastrointestinal, respiratory, cardiovascular, genitourinary, musculoskeletal, endocrine, hematologic, dermatologic, psychiatric, thyroid, coronary, prostate.\\
\emph{Infections:} infection, sepsis, pneumonia, uti, cellulitis, meningitis.\\
\emph{Pathophysiology:} failure, infarction, ischemia, hemorrhage, thrombosis, embolism, obstruction, perforation, rupture, stenosis, regurgitation, hypertrophy, atrophy, neoplasm, malignancy, metastasis, fibrillation, arrhythmia.\\
\emph{Labs/findings:} elevated, decreased, anemia, leukocytosis, thrombocytopenia, hyperglycemia, hypoglycemia, acidosis, alkalosis, hypoxemia, creatinine, bilirubin, troponin, bnp, lactate, wbc, cultures, infiltrate, consolidation, effusion, cardiomegaly, a1c, bmi, cholesterol, lipid, sodium, hyponatremia, ejection, iron, ferritin.\\
\emph{Imaging/procedures:} ultrasound, ct, mri, xray, echo, ekg, stent, cabg.\\
\emph{Treatments:} antibiotics, diuretics, vasopressors, insulin, anticoagulation, oxygen, ventilation, dialysis, transfusion, surgery, metformin, statin, inhaler, cpap, aspirin, opioid, ppi.\\
\emph{Chronic conditions and care factors:} diabetes, diabetic, obesity, obese, copd, asthma, depression, anxiety, apnea, insomnia, sleep, smoking, tobacco, nicotine, ckd, gout, stroke, tia, palliative, hospice, comfort, dnr.
\end{quote}

\section{Paired Bootstrap Procedure and Results}
\label{app:bootstrap}

To distinguish point-estimate differences from sampling noise, we perform a paired bootstrap test on Macro-F1, comparing \shifamind{} to each of the baseline models. For each pairwise comparison, we run $B = 1{,}000$ paired bootstrap replicates using a fixed random seed (seed $= 42$). Each replicate $b$ samples $n$ test indices $\mathcal{I}^{(b)} \subset \{1,\dots,n\}$ uniformly with replacement (where $n = 17{,}088$ is the test-set size), then computes
\[
  \Delta^{(b)} = \mathrm{F1}_{\text{Macro}}(\hat{\mathbf{y}}^{\,\shifamind{}}_{\mathcal{I}^{(b)}}, \mathbf{y}_{\mathcal{I}^{(b)}}) - \mathrm{F1}_{\text{Macro}}(\hat{\mathbf{y}}^{\,\text{other}}_{\mathcal{I}^{(b)}}, \mathbf{y}_{\mathcal{I}^{(b)}}),
\]
where the same indices $\mathcal{I}^{(b)}$ are used for both predictions (paired). Macro-F1 uses the same global threshold $\tau = 0.5$ and zero-division handling as the main-text computation (\Cref{sec:threshold}). The 95\% confidence interval reported in \Cref{tab:bootstrap} is $[\Delta^{(2.5)}, \Delta^{(97.5)}]$, the $2.5$ and $97.5$ percentiles of $\{\Delta^{(b)}\}_{b=1}^{B}$. The two-sided $p$-value is $p = 2\,\min\{\Pr[\Delta^{(b)} \leq 0],\Pr[\Delta^{(b)} \geq 0]\}$, capped at 1. We use the percentile interval rather than a parametric (e.g.\ Gaussian) approximation because the per-replicate $\Delta$ distributions are not symmetric for the largest gaps (e.g.\ Vanilla CBM).

\begin{table}[h]
\centering
\small
\caption{Paired bootstrap test on Macro-F1: \shifamind{} vs.\ every other model. $\Delta$ is the point-estimate difference (\shifamind{} minus other). The 95\% confidence interval is the $[2.5, 97.5]$ percentile of $\Delta$ over 1{,}000 paired bootstrap replicates.}
\label{tab:bootstrap}
\begin{tabular}{l c c c c c}
\toprule
\textbf{Comparison} & \textbf{F1\textsubscript{Ours}} & \textbf{F1\textsubscript{Other}} & \textbf{$\Delta$} & \textbf{95\% CI} & \textbf{$p$-value} \\
\midrule
\shifamind{} vs.\ LAAT          & 0.712 & 0.711 & $+0.001$ & $[-0.001, +0.003]$ & $0.452$ \\
\shifamind{} vs.\ KEPT          & 0.712 & 0.687 & $+0.025$ & $[+0.022, +0.028]$ & $< 10^{-4}$ \\
\shifamind{} vs.\ CAML          & 0.712 & 0.674 & $+0.038$ & $[+0.036, +0.041]$ & $< 10^{-4}$ \\
\shifamind{} vs.\ PLM-ICD       & 0.712 & 0.650 & $+0.062$ & $[+0.059, +0.065]$ & $< 10^{-4}$ \\
\shifamind{} vs.\ GKI-ICD       & 0.712 & 0.649 & $+0.064$ & $[+0.060, +0.067]$ & $< 10^{-4}$ \\
\shifamind{} vs.\ Vanilla CBM   & 0.712 & 0.164 & $+0.548$ & $[+0.545, +0.552]$ & $< 10^{-4}$ \\
\bottomrule
\end{tabular}
\end{table}

\shifamind{} is statistically indistinguishable from LAAT on Macro-F1: the 95\% interval for $\Delta$ contains zero and the two-sided $p$-value is $0.452$. Against the other four ICD-coding baselines and against the capacity-matched Vanilla CBM, the gap is significant at every conventional threshold ($p < 10^{-4}$ in each case), with confidence intervals strictly bounded above zero. 

\shifamind{} matches the strongest existing baseline on point accuracy while being the only model in the comparison that produces concept-mediated explanations.

\textbf{Ranking Metric Bootstrap Results.} We applied the same paired bootstrap procedure to AUC-ROC, P@$K$, and R@$K$ for $K \in \{5, 8, 15\}$. \shifamind{} statistically outperforms CAML, PLM-ICD, KEPT, GKI-ICD, and Vanilla CBM on every ranking metric ($p < 0.01$ in all cases), with $\Delta$ ranging from $+0.0013$ (vs.\ KEPT, Mac-AUC) to $+0.3170$ (vs.\ Vanilla CBM, R@8). Against LAAT, the bootstrap shows small but statistically significant LAAT advantages on ranking quality (Mac-AUC $\Delta = -0.0045$, P@5 $\Delta = -0.0070$), consistent with the precision-favoring calibration reported in \Cref{app:precision_recall}.

\section{Long-Tail Bin Definitions}
\label{app:longtail_bins}

\begin{figure}[h]
\centering
\includegraphics[width=0.9\linewidth]{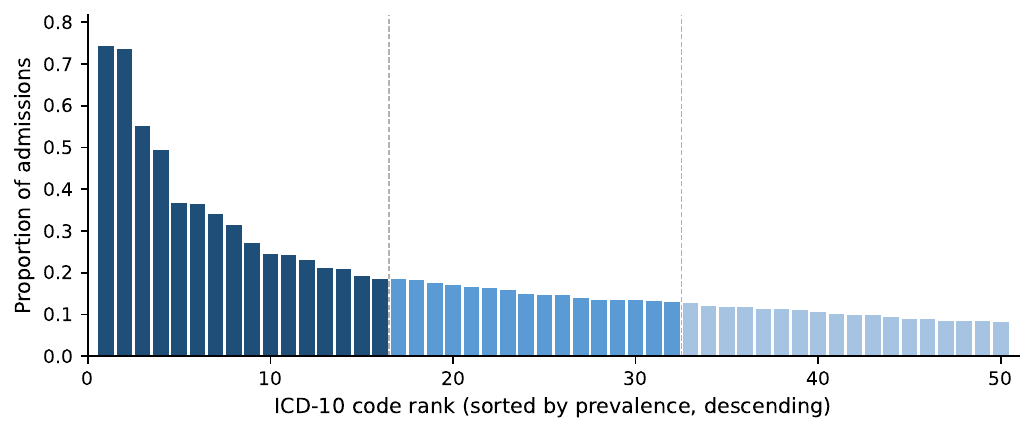}
\caption{Distribution of ICD-10 code prevalence across the 50-code MIMIC-IV top-50 set, sorted by descending prevalence. Each bar shows the proportion of the $113{,}918$ admissions in which the corresponding code is positive. Colors indicate the HEAD/MID/TAIL bins used in \Cref{tab:longtail}. The head-to-tail prevalence ratio is $\sim 9\times$ (E785 hyperlipidemia at $74\%$ vs.\ D72829 elevated WBC at $8\%$), and the curve continues to fall sharply at the TAIL boundary, motivating the rare-code stratified analysis in \Cref{tab:longtail}.}
\label{fig:code_distribution}
\end{figure}

\Cref{fig:code_distribution} presents the long-tail distribution of ICD-10 codes in our dataset. The 50 ICD-10 codes used in \Cref{tab:longtail} are stratified by positive admission count (\Cref{app:icd_codes}) into three bins. Sorting in descending order by admission count, the first 16 codes form HEAD, the next 16 form MID, and the remaining 18 form TAIL. Boundary codes by admission count: HEAD $\geq 21{,}137$ (smallest is D649); MID $\in [14{,}786, 21{,}022]$ (largest is I4891, smallest is G8929); TAIL $\leq 14{,}481$ (largest is Z955, smallest is D72829 at $9{,}435$).

\begin{table}[t]
\centering
\small
\caption{Macro-F1 stratified by code frequency. Codes are sorted by positive admission count and split into HEAD (16 codes), MID (16 codes), and TAIL (18 codes). \textbf{Bold}: best in column. Bin-membership decisions are listed in \Cref{app:longtail_bins}.}
\label{tab:longtail}
\begin{tabular}{l c c c c}
\toprule
\textbf{Model} & \textbf{HEAD} & \textbf{MID} & \textbf{TAIL} & \textbf{Overall} \\
\midrule
CAML                & 0.711           & 0.671           & 0.644           & 0.674 \\
LAAT                & 0.740           & 0.715           & \textbf{0.682}  & 0.711 \\
PLM-ICD             & 0.699           & 0.658           & 0.600           & 0.650 \\
KEPT                & 0.722           & 0.699           & 0.645           & 0.687 \\
GKI-ICD             & 0.696           & 0.658           & 0.598           & 0.649 \\
Vanilla CBM         & 0.341           & 0.124           & 0.042           & 0.164 \\
\midrule
\shifamind{} (Ours) & \textbf{0.742}  & \textbf{0.722}  & 0.677           & \textbf{0.712}\\
\bottomrule
\end{tabular}
\end{table}

\section{Precision/Recall Decomposition at $\tau=0.5$}
\label{app:precision_recall}
Under the fixed threshold $\tau=0.5$, LAAT achieves the highest macro- and micro-precision ($0.757$, $0.778$), whereas \shifamind{} achieves the highest macro- and micro-recall ($0.761$, $0.800$). This suggests that the two models operate at different points under the shared decision rule. We do not interpret this as a calibrated operating-point comparison, but report it to clarify how the Macro-F1 results decompose into precision and recall (\Cref{tab:precision_recall}).
\begin{table}[t]
\centering
\small
\caption{Precision and recall breakdown at $\tau = 0.5$. \textbf{Bold}: best in column. \underline{Underlined}: second best. ShifaMind operates at a recall-favoring point on the same threshold; LAAT at a precision-favoring point.}
\label{tab:precision_recall}
\begin{tabular}{l cccc}
\toprule
\textbf{Model} & \textbf{Macro-Precision} & \textbf{Macro-Recall} & \textbf{Micro-Precision} & \textbf{Micro-Recall} \\
\midrule
CAML \cite{mullenbach2018caml}        & 0.725              & 0.653              & 0.748              & 0.686 \\
LAAT \cite{vu2020laat}                & \textbf{0.757}     & \underline{0.683}  & \textbf{0.778}     & \underline{0.716} \\
PLM-ICD \cite{huang2022plmicd}        & 0.710              & 0.615              & 0.746              & 0.657 \\
KEPT \cite{yang2022kept}              & \underline{0.734}  & 0.658              & \underline{0.761}  & 0.696 \\
GKI-ICD \cite{zhang2025gkiicd}        & 0.719              & 0.610              & 0.752              & 0.654 \\
Vanilla CBM \cite{koh2020cbm}         & 0.348              & 0.162              & 0.415              & 0.338 \\
\midrule
\shifamind{}                          & 0.674              & \textbf{0.761}     & 0.680              & \textbf{0.800} \\
\bottomrule
\end{tabular}
\end{table}

\section{Baseline Implementation Details}
\label{app:baseline_details}

We re-implement each baseline following the architecture and training settings described in the corresponding paper. All baselines are trained on the identical MIMIC-IV top-50 split used by \shifamind{} ($79{,}742$ train, $17{,}088$ validation, $17{,}088$ test admissions; seed $42$), use the same focal or BCE diagnosis loss as in their original papers, and are evaluated at the shared global threshold $\tau = 0.5$ (\Cref{sec:threshold}). Per-baseline configurations are documented below.

\textbf{CAML} \cite{mullenbach2018caml}. We train 100-dimensional Word2Vec embeddings (skip-gram) on the training-set discharge notes and apply a Conv1D layer with kernel size 4, 500 filters, and $\tanh$ activation, followed by a per-label attention head. We use a dropout rate of $0.5$, optimizer Adam with learning rate $10^{-3}$, and batch size $16$.

\textbf{LAAT} \cite{vu2020laat}. The model uses the same Word2Vec embeddings as CAML, encoded by a bidirectional LSTM with hidden size $512$ per direction (yielding a $1{,}024$-dimensional representation), followed by label-aware attention with attention dimension $d_a = 512$. We use a dropout rate of $0.3$, optimizer AdamW with learning rate $10^{-3}$, batch size $8$, and a \texttt{ReduceLROnPlateau} schedule (factor $0.9$, patience $2$).

\textbf{PLM-ICD} \cite{huang2022plmicd}. We use the \texttt{biomed\_roberta\_base} encoder with a chunked-input strategy: each discharge note is split into segments of $128$ tokens, with up to $24$ chunks per note (effective context $3{,}072$ tokens). A LAAT-style label-aware attention head operates over the concatenated chunk representations. We optimize with AdamW at learning rate $5 \times 10^{-5}$, $2{,}000$ warmup steps, and batch size $8$.

\textbf{KEPT} \cite{yang2022kept}. We use the \texttt{whaleloops/keptlongformer} backbone (Longformer pretrained with ICD description and UMLS objectives), which supports a $4{,}096$-token context. The input is constructed by concatenating $50$ \texttt{[description][MASK][SEP]} prompts (one per ICD code) with the discharge note. Global attention is applied to the CLS token and to the $50$ \texttt{[MASK]} positions, and a shared $\mathrm{Linear}(768 \to 1)$ projection is applied to each \texttt{[MASK]} representation to produce per-code logits. We optimize with AdamW at learning rate $1.5 \times 10^{-5}$, weight decay $10^{-3}$, and $\epsilon = 10^{-7}$.

\textbf{GKI-ICD} \cite{zhang2025gkiicd}. The model uses \texttt{biomed\_roberta\_base} with the same chunked-input strategy as PLM-ICD ($128 \times 24$ tokens). A PLM-CA decoder applies cross-attention over $50$ learnable code queries, initialized from the max-pooled embeddings of each ICD code description. The training loss combines four terms: $\mathcal{L}_{\mathrm{raw}} + \alpha\,\mathcal{L}_{\mathrm{R\text{-}Drop}} + \mathcal{L}_{\mathrm{guide}} + \lambda\,\mathcal{L}_{\mathrm{sim}}$, with $\alpha = 10$ and $\lambda = 0.5$. At inference time, only the raw discharge-note pathway is used. We optimize with Adam at learning rate $5 \times 10^{-5}$.

\textbf{Vanilla CBM} \cite{koh2020cbm}. The Vanilla CBM baseline is capacity-matched to \shifamind{}: it uses the same BioClinical-ModernBERT-base encoder, the same $6{,}144$-token context length, the same AdamW optimizer with learning rate $2 \times 10^{-5}$, the same focal diagnosis loss, and the same training schedule. The only architectural difference is the bottleneck. Vanilla CBM applies a strict additive scalar bottleneck,
\[
\hat{\boldsymbol{\ell}} = \mathbf{W}_d\,\sigma(\mathbf{W}_c\,\mathbf{p_t} + \mathbf{b}_c) + \mathbf{b}_d,
\]
in place of the multiplicative gate over the concept-grounded representation $\mathbf{p_c}$ used by \shifamind{}. This isolates the bottleneck form as the sole source of performance differences between the two models.

\section{Analytical Jacobian for CIM}
\label{app:jacobian}

The Concept Influence Magnitude (CIM) metric in \Cref{sec:xai} measures the gradient norm of each diagnosis logit with respect to the representation feeding the diagnosis head, averaged over examples where the relevant concept and diagnosis are jointly positive. We compute these gradients in closed form rather than via automatic differentiation, both to avoid numerical artifacts associated with mixed-precision autograd and to enable batched evaluation on GPU.

Throughout this appendix, $\sigma'(\cdot)$ denotes the element-wise derivative of the sigmoid; $\mathbb{1}[\cdot]$ the element-wise indicator function; $\mathrm{diag}(\mathbf{v})$ the diagonal matrix formed from a vector $\mathbf{v}$; and $\gamma_{\mathrm{LN}} \in \mathbb{R}^{h}$ the LayerNorm scale parameter applied to the gated representation. The gradient $\nabla_{\mathbf{r}}\hat{\ell}_j$ denotes the gradient of the $j$-th diagnosis logit with respect to the representation $\mathbf{r}$ at the input to the diagnosis head ($\mathbf{p_c}$ for \shifamind{}'s MCB; $\hat{\mathbf{c}}$ for the Vanilla CBM, abbreviated VCBM).

\paragraph{Vanilla CBM.} For VCBM, the diagnosis pathway is $\mathbf{p_t} \to \hat{\mathbf{c}} = \sigma(\mathbf{W}_c \mathbf{p_t} + \mathbf{b}_c) \to \hat{\boldsymbol{\ell}} = \mathbf{W}_d\,\hat{\mathbf{c}} + \mathbf{b}_d$. The literal input to the diagnosis head is $\hat{\mathbf{c}} \in \mathbb{R}^C$, and the gradient is closed-form and constant in the input:
\[
\nabla_{\hat{\mathbf{c}}}\hat{\ell}_j = \mathbf{W}_d[j,:].
\]
Hence $\mathrm{CIM}_{c,j} = \|\mathbf{W}_d[j,:]\|_2$ for any $(c,j)$ pair with at least one co-positive sample, and the aggregate macro-CIM is the mean over valid pairs.
\paragraph{Sensitivity at a common-dimensional surface.} The literal input to each model's diagnosis head differs in dimensionality ($\mathbf{p_c} \in \mathbb{R}^{768}$ for \shifamind{}, $\hat{\mathbf{c}} \in \mathbb{R}^{160}$ for VCBM). As a complementary, dimension-controlled comparison, we also measure CIM at the encoder output $\mathbf{p_t} \in \mathbb{R}^{768}$ for both models. At this surface, $\nabla_{\mathbf{p_t}} \hat{\ell}_j = \mathbf{W}_d[j,:]\,\mathrm{diag}(\sigma'(\mathbf{W}_c \mathbf{p_t} + \mathbf{b}_c))\,\mathbf{W}_c$ for VCBM (chain rule through the sigmoid bottleneck), and $\nabla_{\mathbf{p_t}}\hat{\ell}_j$ for \shifamind{} passes through the gate-network input. Aggregate CIM at $\mathbf{p_t}$: \shifamind{} $1.314$ vs.\ VCBM $0.034$, a ratio of approximately $39\times$. The two formulations answer different questions: at the diagnosis-head input ($2.0\times$ ratio) we compare the gradient surfaces the head literally consumes; at $\mathbf{p_t}$ ($39\times$ ratio) we compare end-to-end encoder-to-logit sensitivity at a common $768$-dimensional surface, which captures sigmoid attenuation along VCBM's bottleneck path.
\paragraph{ShifaMind MCB.} For \shifamind{}, the gradient passes through the gate network and the LayerNorm. We define the intermediate quantities
\begin{align*}
\mathbf{h}_1 &= \mathbf{W}_1\,[\mathbf{p_t};\mathbf{p_c}] + \mathbf{b}_1, &
\mathbf{h}_2 &= \mathbf{W}_2\,\relu(\mathbf{h}_1) + \mathbf{b}_2, \\
\mathbf{g} &= \sigma(\mathbf{h}_2), &
\mathbf{u} &= \mathbf{g}\odot\mathbf{p_c}, \\
\mathbf{z} &= \mathrm{LayerNorm}(\mathbf{u}).
\end{align*}
The diagnosis logit decomposes as $\hat{\ell}_j = \mathbf{W}_d[j,:]\,\mathbf{z} + b_{d,j}$, and the gradient with respect to $\mathbf{p_c}$ is the product of three Jacobians: from $\mathbf{p_c}$ to the gate $\mathbf{g}$, from $\mathbf{p_c}$ through the element-wise product $\mathbf{u}$, and from $\mathbf{u}$ through the LayerNorm to $\mathbf{z}$:
\begin{align}
J_{\mathbf{g}\to\mathbf{p_c}} &= \mathrm{diag}\!\left(\sigma'(\mathbf{h}_2)\right)\mathbf{W}_2\,\mathrm{diag}\!\left(\mathbb{1}[\mathbf{h}_1 > 0]\right)\mathbf{W}_{1,\mathbf{p_c}}, \\
J_{\mathbf{u}\to\mathbf{p_c}} &= \mathrm{diag}(\mathbf{g}) + \mathrm{diag}(\mathbf{p_c})\,J_{\mathbf{g}\to\mathbf{p_c}}, \\
J_{\mathbf{z}\to\mathbf{u}} &= \tfrac{1}{s}\,\mathrm{diag}(\gamma_{\mathrm{LN}})\!\left(\mathbf{I} - \tfrac{1}{h}\,\mathbf{1}\mathbf{1}^\top - \tfrac{1}{h}\,\mathbf{f}\mathbf{f}^\top\right),
\end{align}
where $\mathbf{W}_{1,\mathbf{p_c}}$ is the second $h$-column block of $\mathbf{W}_1$ corresponding to the $\mathbf{p_c}$ portion of the concatenated input, $s$ is the per-sample standard deviation of $\mathbf{u}$, and $\mathbf{f} = (\mathbf{u} - \mathrm{mean}(\mathbf{u}))/s$ is the standardized residual. Composing these factors yields the closed-form gradient
\[
\nabla_{\mathbf{p_c}}\hat{\ell}_j = \mathbf{W}_d[j,:]\,J_{\mathbf{z}\to\mathbf{u}}\,J_{\mathbf{u}\to\mathbf{p_c}}.
\]
The norm $\|\nabla_{\mathbf{p_c}}\hat{\ell}_j\|_2$ used in CIM is computed in batched form on GPU; matrix products with diagonal factors are implemented as element-wise broadcasts to avoid materializing the full diagonal matrices.

\section{Training Details}
\label{app:training}

\shifamind{} is trained with AdamW ($\beta_1 = 0.9$, $\beta_2 = 0.999$, $\epsilon = 10^{-8}$) at a learning rate of $2 \times 10^{-5}$ with a linear schedule and 10\% warmup steps. We train for 5 epochs with batch size 8, using bfloat16 mixed precision on a single NVIDIA A100 90GB GPU. The best checkpoint is selected by validation Macro-F1. Hyperparameters were chosen based on standard fine-tuning settings for transformer-based clinical NLP models and validation-set performance. All other hyperparameters are as described in \Cref{sec:method}.

\textbf{Wall-clock runtime.} Training times are reported as approximate ranges; we do not report exact wall-clock measurements because runs were not instrumented with precise timing logs. On the same MIMIC-IV top-50 split and a single A100 90GB GPU, \shifamind{} takes approximately 11--12 hours to train end-to-end. Baseline runtimes under matched conditions are approximately 8 hours for CAML; 10--12 hours each for LAAT, PLM-ICD, GKI-ICD, and Vanilla CBM; and 14 hours for KEPT (driven by its $4{,}096$-token prompt-augmented input).

\section{NegEx Implementation}
\label{app:negex}

Our negation detector follows Chapman et al.\ \cite{chapman2001negex} with pre-negation triggers (\emph{no, not, without, denies, absent, negative for, no evidence of, free of, rules out, \dots}), post-negation triggers (\emph{was ruled out, were negative, not present, \dots}), and pseudo-negation triggers (\emph{not only, no increase, without difficulty, \dots}). For each concept occurrence, we extract a six-token scope on either side, truncated at sentence boundaries or contrastive conjunctions (\emph{but, however, although, yet}). A concept is marked positive only if at least one occurrence falls outside all negation scopes. On a 5{,}000-note sample, NegEx reduces naive keyword activations from 137{,}602 to 122{,}762 (10.8\% correction rate).

\section{Concept-Mask Intervention: Implementation Details}
\label{app:concept_mask}

\textbf{TopC mapping.} For each diagnosis $j$, $\mathrm{TopC}(j)$ is the five concepts with highest Pearson correlation between concept presence (NegEx-derived training labels) and diagnosis label on the training set. This is the same mapping used for the CSTPR metric in \Cref{sec:xai}. Examples: $\mathrm{TopC}(\text{E785}) = \{\text{aspirin}, \text{hypertension}, \text{coronary}, \text{cabg}, \text{diabetes}\}$; $\mathrm{TopC}(\text{I10}) = \{\text{hypertension}, \text{metformin}, \text{aspirin}, \text{surgery}, \text{cholesterol}\}$.

\textbf{Span matching.} For each (note, target diagnosis $j$), we identify token positions corresponding to any $c \in \mathrm{TopC}(j)$ via case-insensitive whole-word regex match (\texttt{\textbackslash b<concept>\textbackslash b}) against the decoded note text, then map matched character spans to token indices using the tokenizer's offset mapping. Negation handling is not applied at intervention time: any lexical occurrence of a TopC concept name is masked regardless of polarity. Matched tokens are replaced with the encoder's mask token.

\textbf{Pair sampling.} We collect all (note $i$, diagnosis $j$) pairs from the test set where $y_{ij} = 1$ and $\hat{y}_{ij} = 1$ at $\tau = 0.5$ ($72{,}305$ pairs). We shuffle (seed 42) and take the first $1{,}000$. Of these, $917$ contain at least one TopC token and are usable; the remaining $83$ are dropped.

\textbf{Drop measurements.} Let $p_j^{(i)}$ and $\tilde{p}_j^{(i)}$ be the model's predicted probability for diagnosis $j$ before and after masking the target's TopC tokens. The targeted drop is $\Delta_j^{(i)} = p_j^{(i)} - \tilde{p}_j^{(i)}$. The within-note control averages $p_{j'}^{(i)} - \tilde{p}_{j'}^{(i)}$ over all other positive diagnoses $j' \neq j$ in the same note ($y_{ij'} = 1$), evaluated on the \emph{same masked input}. This isolates target-specific sensitivity from general perturbation effects of mask-token insertion. Of the $917$ valid pairs, $18$ ($2.0\%$) have no other positive diagnosis in the note; for these, the control drop is set to zero. A sensitivity analysis restricted to the $899$ pairs with a within-note control yields nearly identical statistics (mean difference $0.113$, $95\%$ CI $[0.100, 0.126]$, sign test $p < 10^{-29}$).

\textbf{Statistics.} Across the $917$ valid pairs, the mean targeted drop is $0.134$, the mean within-note control drop is $0.020$, and the mean per-pair difference is $0.114$. The bootstrap $95\%$ CI of the mean difference is $[0.103, 0.127]$ over $1{,}000$ replicates (seed 42). The targeted drop exceeds the control in $631/917$ pairs ($68.8\%$); a two-sided binomial sign test against $p_0 = 0.5$ yields $p = 1.3 \times 10^{-30}$.

\textbf{Limitations of this test.} Span matching is purely lexical: it may miss paraphrastic mentions of a concept and may include false positives where a concept word appears in unrelated context. The within-note control isolates target-specific sensitivity but does not rule out shared evidence between target and control diagnoses; the partial control drop of $0.020$ is consistent with this. We treat this experiment as behavioral evidence of concept sensitivity rather than proof of one-to-one concept-to-diagnosis attribution; this caveat is noted in \Cref{sec:limitations}.

\end{document}